\documentclass[letterpaper]{article} 
\usepackage{aaai24}  
\usepackage{times}  
\usepackage{helvet}  
\usepackage{courier}  
\usepackage[hyphens]{url}  
\usepackage{graphicx} 
\usepackage{natbib}  
\usepackage{caption} 
\usepackage{algorithm}
\usepackage{algorithmic}
\usepackage{amsmath}

\usepackage{amsfonts,amssymb}
\usepackage{bm}
\usepackage{color}

\DeclareMathOperator*{\argmin}{arg\,min}
\usepackage{booktabs}
\usepackage{multirow}
\usepackage{makecell}

\usepackage{newfloat}
\usepackage{listings}
\DeclareCaptionStyle{ruled}{labelfont=normalfont,labelsep=colon,strut=off} 
\floatstyle{ruled}
\newfloat{listing}{tb}{lst}{}
\floatname{listing}{Listing}
\nocopyright
\title{Align before Search: Aligning Ads Image to Text for Accurate Cross-Modal Sponsored Search}

\author {
    Yuanmin Tang\textsuperscript{\rm 1,\rm 2},
    Jing Yu\textsuperscript{\rm 1,\rm 2}\thanks{Corresponding author},
    Keke Gai\textsuperscript{\rm 3}, 
    Yujing Wang\textsuperscript{\rm 4},
    Yue Hu\textsuperscript{\rm 1}, 
    Gang Xiong\textsuperscript{\rm 1},
    Qi Wu\textsuperscript{\rm 5}
}
\affiliations {
    \textsuperscript{\rm 1}Institute of Information Engineering, Chinese Academy of Sciences \\
    \textsuperscript{\rm 2}School of Cyber Security, University of Chinese Academy of Sciences \\
    \textsuperscript{\rm 3}Beijing Institute of Technology \\
    \textsuperscript{\rm 4}Microsoft Research Asia \\
    \textsuperscript{\rm 5}University of Adelaide \\
\{tangyuanmin, yujing02, huyue, xionggang\}@iie.ac.cn, gaikeke@bit.edu.cn, yujwang@microsoft.com, qi.wu01@adelaide.edu.au
}

\usepackage{bibentry}

\begin{document}

\maketitle

\begin{abstract}
Cross-Modal sponsored search displays multi-modal advertisements (ads) when consumers look for desired products by natural language queries in search engines. Since multi-modal ads bring complementary details for query-ads matching, the ability to align ads-specific information in both images and texts is crucial for accurate and flexible sponsored search. Conventional research mainly studies from the view of modeling the implicit correlations between images and texts for query-ads matching, ignoring the alignment of detailed product information and resulting in suboptimal search performance.In this work,  we propose a simple alignment network for explicitly mapping fine-grained visual parts in ads images to the corresponding text, which leverages the co-occurrence structure consistency between vision and language spaces without requiring expensive labeled training data. Moreover, we propose a novel model for cross-modal sponsored search that effectively conducts the cross-modal alignment and query-ads matching in two separate processes. In this way, the model matches the multi-modal input in the same language space, resulting in a superior performance with merely half of the training data. Our model outperforms the state-of-the-art models by 2.57\%  on a large commercial dataset. Besides sponsored search, our alignment method is applicable for general cross-modal search. We study a typical cross-modal retrieval task on the MSCOCO dataset, which achieves consistent performance improvement and proves the generalization ability of our method. Our code is available at https://github.com/Pter61/AlignCMSS/.
\end{abstract} 

\section{1. Introduction}
\label{sec:intro}

\begin{figure}[t]
    \setlength{\belowcaptionskip}{-5pt}
    \centering
    \includegraphics[width=1.0\linewidth]{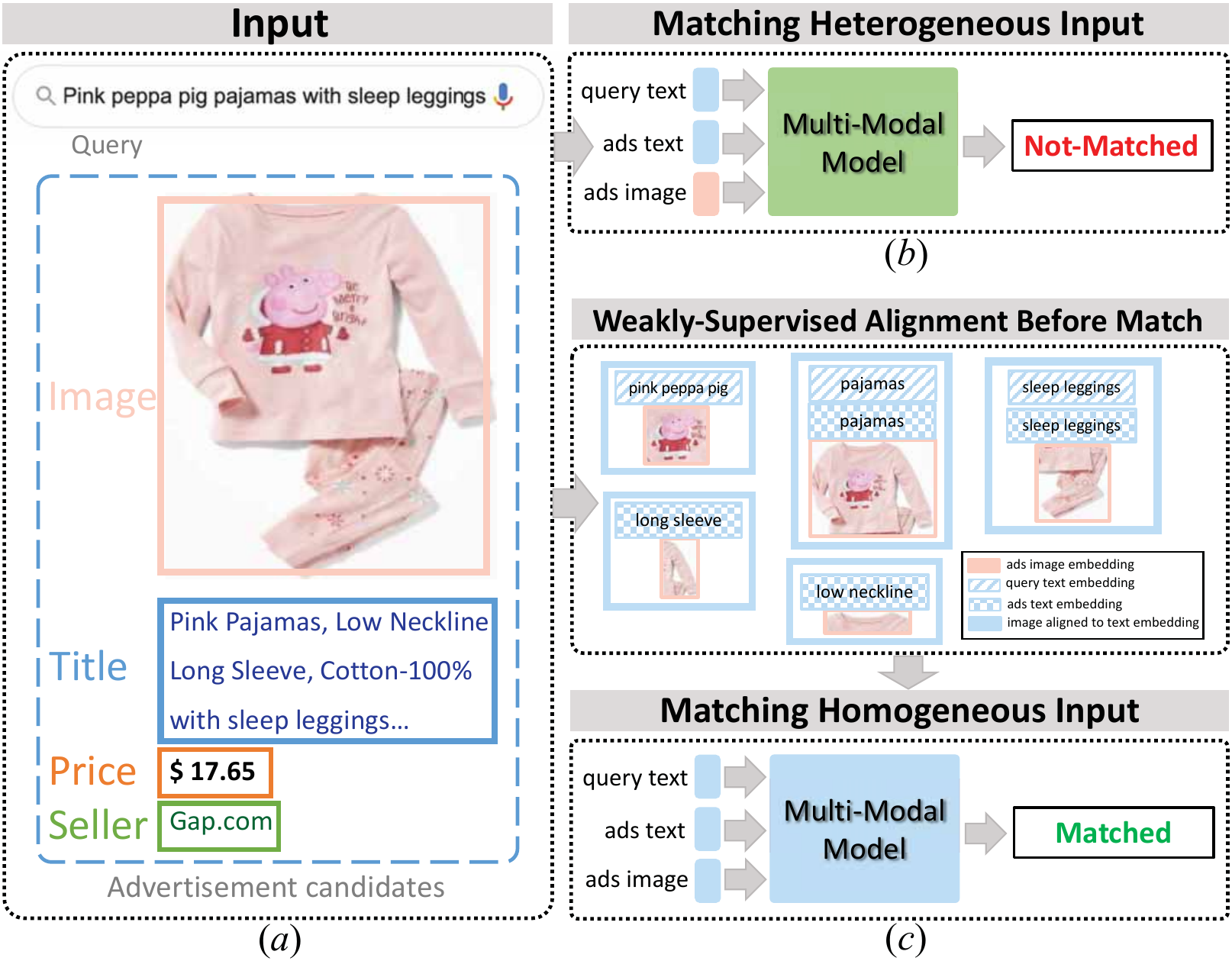}
    \caption{Illustration of our motivation. (b) Existing model with heterogeneous input (a). (c) Our model with homogeneous input is aligned by our weakly-supervised method.} 
    \label{fig:motivation}
\end{figure}

Sponsored search \cite{jansen2008sponsored, ling2017model, 10.1145/3503161.3548226} is a widely used business model on search engine platforms, where sponsored ads are presented to users with other search results. Displaying the relevant ads for a query will significantly increase the exposure of proper products while effectively satisfying the buyers' demand. Thus, the sponsored search system needs to model the relevance between queries and ads accurately.  Besides ads with pure text, multi-modal ads with images have become a new trend and extend the current single-modal sponsored search to Cross-Modal Sponsored Search (CMSS). Images in ads bring complementary product details to texts. Considering the query in Figure \ref{fig:motivation}(a), the agent not only needs to semantically match the ads text of ``pink pajamas'' and ``sleep leggings'', but also has to align the query to the visual content of ``pink peppa pig'' in the ads image. Therefore, accurately capturing the query-relevant content from both visual and textual ads information and jointly matching it with the query is essential to achieve flexible sponsored search.

Recent works \cite{10.1145/3503161.3548226} regard the cross-modal sponsored search as a traditional image-text matching task and leverage image-text correlation learning approaches to model the relevance between the text query and multi-modal ads. The typical solution, i.e., AdsCVLR \cite{10.1145/3503161.3548226}, is based on the vision and language pre-training (VLP) framework, which treats the query and ads text together as text input and regards the corresponding ads image as visual input. All the inputs are fed into a single-stream VLP model to measure their global similarity as illustrated in Figure \ref{fig:motivation}(b). However, this kind of solution is hard to satisfy fine-grained sponsored search because of the great difference between query-ads matching and image-text matching. In sponsored search, the multi-modal ads contain more detailed product information, \textit{e.g.}, color, brand, style, price, and seller. The queries are short \cite{10.1145/3503161.3548226} and generally about partial features of the products (\textit{e.g.}, the query in Figure \ref{fig:motivation}(a) focuses on  attribute ``pink peppa pig''). Thus, aligning fine-grained visual regions in ads images to the corresponding text becomes a core issue for high-quality CMSS. Recent works, which implicitly correlate visual regions and words by attention mechanism in VLP, are challenging to accurately achieve fine-grained alignment without expensive region-word labels for training (empirical evidence is provided in Figure \ref{fig:t-SNE} and supplementary material). 
Moreover, existing works jointly learn cross-modal alignment and query-ads matching by unified transformers, which is data-exhaustive to achieve optimal performance.  

To tackle these issues, we propose a new cross-modal sponsored search model that successively conducts cross-modal alignment and query-ads matching by two independent modules. As shown in Figure \ref{fig:motivation}(c), in the cross-modal alignment module, the visual part embeddings of ads images are aligned to the corresponding semantic word embeddings by a weakly-supervised approach without extra labels. We name this module VALSE for short. Specifically, VALSE learns a simple and effective linear mapping for vision-to-language alignment. As illustrated in Figure \ref{fig:motivation_2}, the alignment approach is based on our observation that the topological structure constructed by the co-occurrence relationships between product partial characteristics is consistent in both vision and language spaces (evidence is provided in the supplementary). Based on this hypothesis, we propose a three-stage alignment approach to progressively map the visual embeddings to the language space by unsupervised adversarial training and weakly-supervised refinement. Experimental results demonstrate the superior alignment quality and generalization ability of our alignment approach. 

In this way, VALSE unifies the multi-modal ads input in the same language space. Then we propose a novel model AlignCMSS for cross-modal sponsored search by combining VALSE with a typical VLP model. AlignCMSS receives the aligned homogeneous embeddings of query, ads text, and ads image from VALSE and matches the query and aligned multi-modal ads in the same language space via the VLP model. By successively conducting cross-modal alignment and query-ads matching, AlignCMSS significantly outperforms existing CMSS models. Moreover, this align-before-match strategy is training-efficient, which  merely uses 50\% training data that results in superior performance compared to the SoTA approaches with the total training data. 

The main contributions are summarized as follows. (1)
We propose VALSE, a novel method for aligning fine-grained visual parts in ads images to the corresponding text without requiring expensive labeled training data of region-word pairs.
VALSE leverages the prior knowledge of co-occurrence structure consistency between vision and language spaces and achieves superior alignment quality for flexible sponsored search. 
(2) We propose AlignCMSS, a novel model for CMSS that effectively conducts the cross-modal alignment and query-ads matching in two separate processes. 
AlignCMSS  demonstrates consistent improvement over various CMSS models and significantly outperforms the state-of-the-art (SoTA) model by 2.57\% on the existing largest commercial dataset. Moreover, AlignCMSS is training-efficient, which sheds new
lights towards the efficient pre-training on large-scale vision-language data. 
(3) Besides sponsored search, the proposed alignment approach is generally applicable to other cross-modal retrieval tasks and has impacts a broader range of applications. We case-study another cross-modal retrieval task on the MSCOCO dataset and achieve a consistent performance improvement of 1.3\% on average precision.

\section{2. Related Work}

\noindent\textbf{Relevance Modeling in Sponsored Search}. Relevance modeling is a critical component of information retrieval. 
Previous works \cite{grbovic2018real, li2021adsgnn, zhu2021textgnn} have focused on language models to extract information from the text while ignoring the value of visual information. FashionBERT \cite{gao2020fashionbert} considers both image and text information to address image-text matching in the fashion domain. AdsCVLR \cite{10.1145/3503161.3548226} first proposes a VLP approach for cross-modal sponsored search to model the relevance of the query, ads text, and ads images, which obtains obvious performance boost compared with existing VLP models. AdsCLVR also trains a two-stream student model using knowledge distillation for online computation and latency constraints. Since the multi-modal inputs are unaligned, it is hard for AdsCLVR to accurately correlate the regions in ads images with words in the query, resulting in incorrect prediction when the ads texts are insufficient or inaccurate.  We propose a cross-modal alignment approach that enables object-level fine-grained vision-language alignment for accurate sponsored search.

\noindent\textbf{Vision and Language Alignment}. Existing methods \cite{zhen2019deep,gao2020fashionbert,yu2018mattnet, Liu_2019_CVPR,anderson2018bottom} typically focused on specific tasks through joint space learning or cross-modal attention. Recent approaches \cite{su2020vl-bert,li2020oscar,zhang2021vinvl,radford2021learning,yao2022filip} aim to develop unified models for cross-modal alignment using vision-language pre-training (VLP). However, VLP models with implicit attention mechanisms struggle to achieve accurate alignment between visual objects and words without fine-grained annotations, which are essential for sponsored search (empirical evidence in supplementary material). Align-before-fuse VLP methods \cite{NEURIPS2021_50525975, pmlr-v162-li22n} attempt to overcome this by employing a tunable patch-based visual encoder and an image-text contrastive loss. Nonetheless, the alignment still relies on global image-text pair supervision, lacking the fine-grained annotations of patch-word pairs necessary for accurate alignment. To address these limitations, we propose a novel alignment method that directly aligns fine-grained image object embeddings with language in a weakly-supervised way. Our alignment stage is independent of relevance modeling, enhancing the cross-modal model (e.g., VLP) better to capture query-ads relevance with the same amount of training data. Furthermore, our method accounts for structure consistency and easily integrates with VLP models, unlike earlier studies \cite{gupta2019vico, wang2020consensus} reliant on concept co-occurrence relationships.

\noindent\textbf{Unsupervised Word Translation}. The study of unsupervised word translation inspires our weakly-supervised solution for vision-language alignment. Existing works \cite{lample2018word,artetxe2019effective,pourdamghani2019translating} use adversarial training to learn a linear mapping between two languages without parallel corpora. In \cite{lample2018word}, a bilingual dictionary is built between two languages without parallel corpora. To extend this schema to cross-modal translation, \cite{chung2018unsupervised} proposes a speech-word-to-language-word alignment method based on adversarial training for speech-to-text translation. However, these previous works only align sequence-structured modalities, which presents difficulties in the vision-language scenario due to the semantic gap. To overcome these limitations, we propose a novel three-stage alignment approach that enables fine-grained alignment between vision and language modalities, effectively addressing the semantic gap between them.

\section{3. VALSE}
\label{sec:VALSE}
The data in the cross-modal sponsored search task is defined as a set of triples: $\mathcal{L}$ = $< \boldsymbol{q}_i, \boldsymbol{a}_i, \boldsymbol{y}_i >$, where $\boldsymbol{q}_i$ denotes the user query, $\boldsymbol{a}_i$ represents a multi-modal ads containing the product image and text, and $\boldsymbol{y}_i\in{\{0, 1\}}$ is the relevance label between $\boldsymbol{q}_i$ and $\boldsymbol{a}_i$, where 0 denotes irrelevant while 1 denotes relevant. This task aims to learn a binary classifier $f : f(\boldsymbol{q}_i, \boldsymbol{a}_i)\in{\{0, 1\}}$ to accurately predict the relevance between user queries and multi-modal ads. 

\begin{figure}[t]
    \centering
    \includegraphics[width=1.0\linewidth]{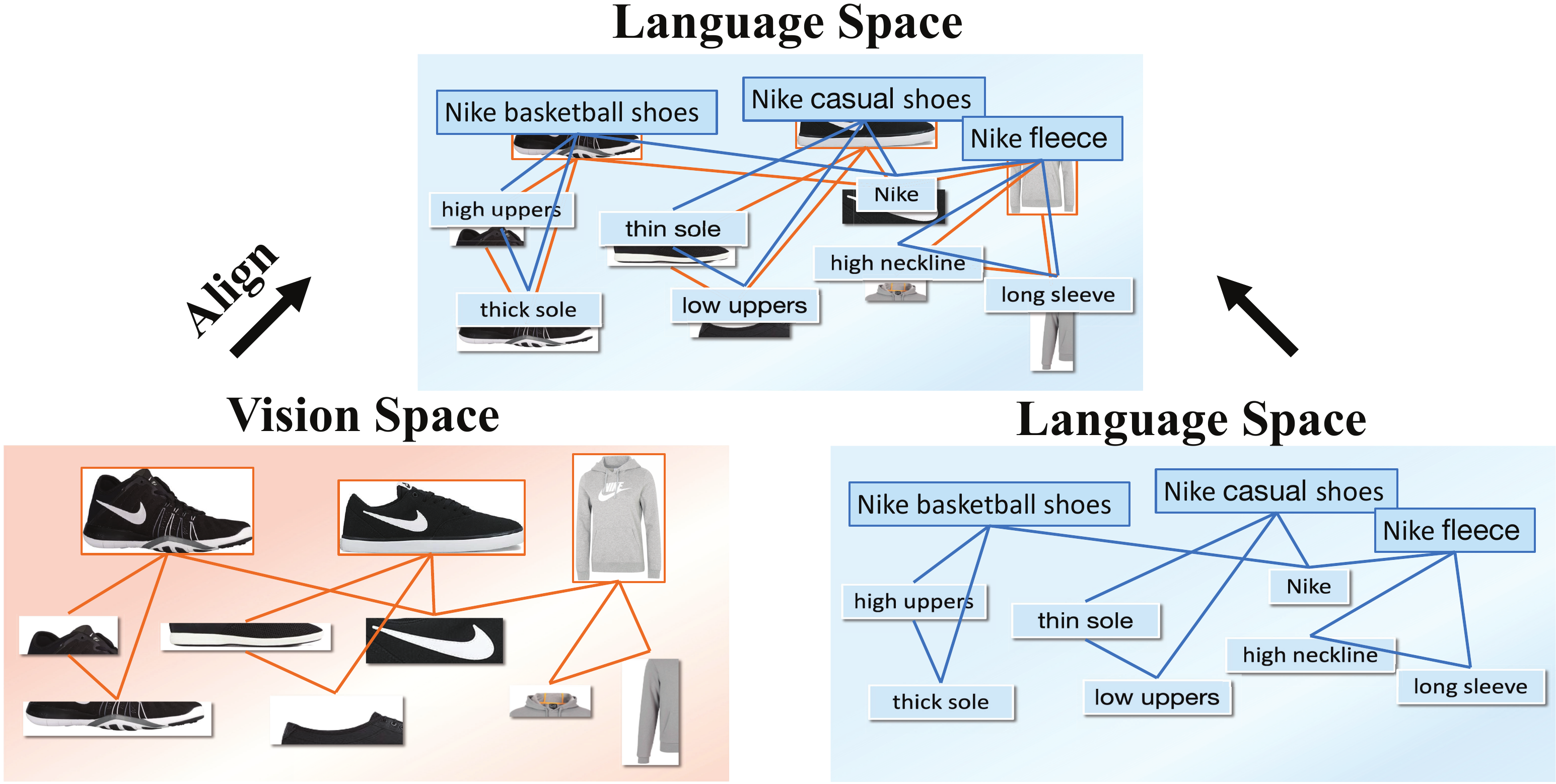}
    \caption{An illustration of structure consistency in vision and language space motivating cross-modal alignment in VALSE. 
    } 
    \label{fig:motivation_2}
\end{figure}

VALSE aims to learn a mapping to align the region embeddings in ads images to the word embeddings with corresponding semantics in ads texts and queries. In this way, multi-modal input embeddings are unified in language space for direct relevance learning in the Vision-Language Pretraning (VLP) model.  The alignment strategy derives from our understanding of the origin of vision and language: both modalities originate from a shared reality.  Therefore, relevant visual components appear in the same scenario while humans talk continuously about relevant things. As shown in Figure \ref{fig:motivation_2}, ``Nike'',  ``thick sole'', and ``high uppers''  have similar co-occurrence relationships for the product ``Nike basketball shoes'' in both ads image and ads text. Thus we consider that the topological structure constructed by co-occurrence relationships between \textit{partial product characteristics} is consistent in both vision and language space. Our empirical study in the supplementary and a paper published in \textit{Nature Machine Intelligence} \cite{roads2020learning} has a similar conclusion. Inspired by this observation, we propose to align ads image regions and corresponding text words by aligning their co-occurrence structure constructed by their embeddings instead of learning the mapping directly from the ground-truth text-image pairs. In this way, we don't require labor-extensive text-image annotations. In this section, we first introduce the vision and language embedding extraction approach to keep the co-occurrence structure in a single modality space. Then we propose a three-stage alignment approach based on the structure consistency across different modalities.

\subsection{3.1. Vision and Language Space Construction}
\label{ssec:space construction}

To keep the co-occurrence structure in each modality space, we propose representing the co-occurrence information in the object (\textit{resp.}, word) embeddings. Specifically, in the learned vision (\textit{resp.}, language) embedding space, objects (\textit{resp.}, words) with similar co-occurrence relationships are close to each other. Such structure-preserving constraints have been extensively studied in metric learning \cite{Wang_2016_CVPR, liu2021semantic}. To make our alignment approach more generic and applicable to existing models, we utilize the widely-used embedding approaches in existing models with implicit co-occurrence learning, \textit{e.g.}, object detector X152-C4 for region embeddings, and transformer for word embeddings. Such implicit co-occurrence embeddings have obtained substantial benefits for cross-modal alignment. Since our work mainly focuses on proving the effectiveness of the align-before-match framework and the alignment strategy, other regularization terms for co-occurrence structure preserving can be further studied.      

\noindent\textbf{Vision Embedding Space Construction.} Since we select VinVL as the baseline, we use the same object detector X152-C4 \cite{zhang2021vinvl} in VinVL to detect ads image regions and extract their embeddings to construct the vision space. In the learning process of the object detector, objects in an image are detected and embedded jointly, which implicitly introduces object co-occurrence constraints in each detected object embedding and preserves weak co-occurrence structure.  We denote the set of detected objects as $\mathcal{O} = \{{\boldsymbol{o}_i}\}_{i=1}^K (K=50)$ and represent each object $\boldsymbol{o}_i$  by a region embedding $\boldsymbol{v}_i\in{\mathbb{R}^{d_1}}$  ($d_1= 2054$ with 2048-dimensional visual embedding and 6-dimensional position embedding). The embeddings extracted from all the ads images in the commercial advertising dataset \cite{10.1145/3503161.3548226} form the vision space. 

\begin{figure*}
    \centering
    \vspace{5pt}
    \includegraphics[width=1.0\linewidth]{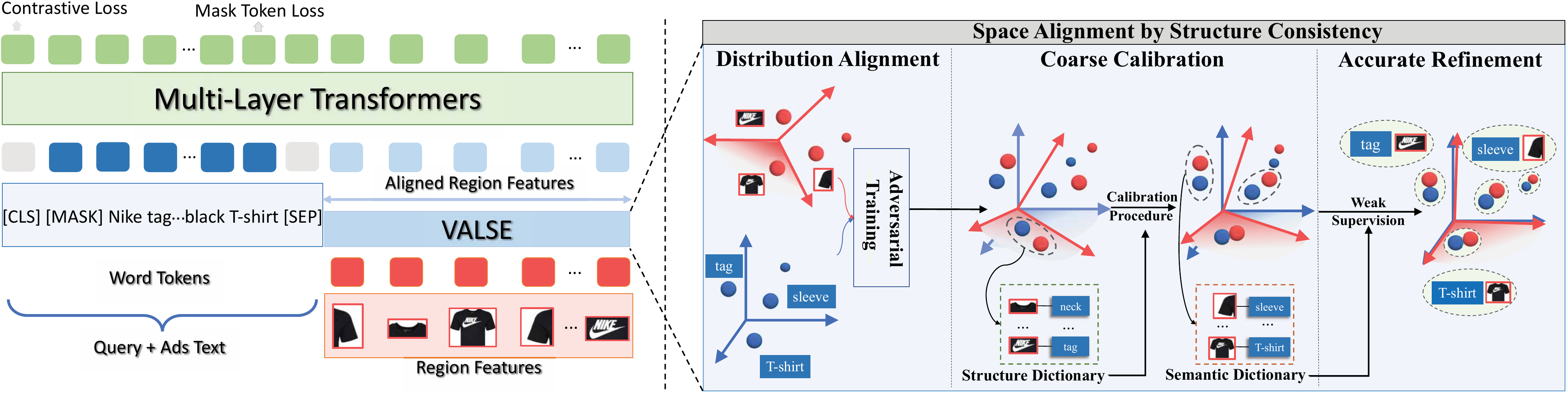}   
    \caption{An overview of AlignCMSS with two parts: (left) the architecture of AlignCMSS and (right) three alignment stages in VALSE. We first train VALSE offline, then replace the original linear mapping in VinVL with VALSE and fine-tune it online. }
 \vspace{-10pt}
    \label{fig:model-architecture}
\end{figure*}

\noindent\textbf{Language Embedding Space Construction.} 
Transformers have been widely verified the ability to model contextual information for word embeddings \cite{devlin2019bert, floridi2020gpt}. Thus the information of co-occurring words is implicitly correlated in the output embeddings of Transformers by the attention mechanism. Therefore, we apply the transformer-based language encoder to extract the word embeddings to construct the language space. Since our backbone VinVL is based on transformer structure,  we utilize the language part in VinVL multi-modal encoder to obtain the output word embeddings for alignment. Specifically, we tokenize the query $\boldsymbol{q}_i$ and ads text in $\boldsymbol{a}_i$  and obtain a sequence of $D$ tokens with BERT embeddings \cite{devlin2019bert}. 
All the word embeddings with padding image region embeddings are fed into VinVL to extract word embeddings $\{\boldsymbol{t}_j\}_{j=1}^D \in{\mathbb{R}^{d_2}}$ ($d_2$=768) from the output layer. We only keep noun embeddings to form the language embedding space for alignment since the image regions are mostly objects, and the noun embeddings contain contextual information about the nouns. 

\subsection{3.2. Space Alignment by Structure Consistency}
\label{ssec:space alignment}
Given a set of $m$ region embeddings $\mathcal{V}$ = \{$\boldsymbol{v}_i\}_{i=1}^m\subseteq{\mathbb{R}^{d_1\times{m}}}$  and a set of $n$ noun embeddings $\mathcal{T}$ = $\{\boldsymbol{t}_j\}_{j=1}^n\subseteq{\mathbb{R}^{d_2\times{n}}}$, we aim to learn a linear mapping $\boldsymbol{W_{{align}}}\in{\mathbb{R}^{d_2\times{d_1}}}$ to align the two spaces as: 
\begin{equation}
\begin{aligned}
\boldsymbol{W_{align}} = \argmin_{\boldsymbol{W}\in{\mathbb{R}^{d_2\times{d_1}}}}(||\boldsymbol{W}\mathcal{V} - \mathcal{T}||^2)
\end{aligned}
\label{f:Walign}
\end{equation}
We use linear mapping instead of complex non-linear mapping for alignment since we align the co-occurrence structures of regions and nouns instead of the one-to-one features of region-noun pairs. The co-occurrence structures of vision and language spaces have natural consistency, which has been proven effective for linear mapping to align in our study and previous bilingual word alignment study \cite{lample2018word}.  In this work, we do not have a golden dictionary that specifies which $\boldsymbol{v}_i$ corresponds to $\boldsymbol{t}_j$. 
We propose a weakly-supervised cross-modal alignment approach with three stages: \textit{Distribution Alignment} first coarsely maps the distribution of region embeddings with the distribution of noun embeddings via adversarial training; \textit{Coarse Calibration} then selects the best mutually aligned region-noun embeddings as a structural dictionary to calibrate $\boldsymbol{W_{{align}}}$;  \textit{Accurate Refinement} automatically constructs region-noun semantic labels for weakly-supervised refinement of $\boldsymbol{W_{{align}}}$.

\noindent\textbf{Distribution Alignment by Adversarial Training.} 
This stage coarsely aligns the distribution of vision and language embeddings by making the mapped region features $\boldsymbol{W_{{align}}}\mathcal{V}$ and noun embeddings $\mathcal{T}$ indistinguishable. Since adversarial training is a strong unsupervised solution to achieve this goal, we apply an adversarial approach for learning $\boldsymbol{W_{{align}}}$ without cross-modal supervision. 
We define the alignment mapping $\boldsymbol{W_{{align}}}$ as the generator and a binary classifier as the discriminator, which is a multilayer perceptron with two hidden layers parameterized by $\theta_D$. The generator aligns region embeddings to the language space by  $\boldsymbol{W_{{align}}}\boldsymbol{v}_i$, while the discriminator distinguishes between the transformed region embeddings and noun embeddings $\boldsymbol{t}_j$ by minimizing the following objective:
\begin{equation}
\resizebox{0.9\hsize}{!}{$\begin{aligned}
\mathcal{L}_D(\theta_D|\boldsymbol{W_{{align}}}) = &-\frac{1}{m}\sum_{i = 1}^m\log{P_{\theta_D}({\rm vis} = 1|\boldsymbol{W_{{align}}}\boldsymbol{v}_i)} \\
                                                    &-\frac{1}{n}\sum_{j = 1}^n\log{P_{\theta_D}({\rm vis} = 0|\boldsymbol{t}_j)} 
\end{aligned}$}
\label{f:2}
\end{equation}
where $P_{\theta_D}({\rm vis} = 1|\boldsymbol{W_{{align}}}\boldsymbol{v}_i)$ denotes the probability that $\boldsymbol{v}_i$ is from the vision space while $P_{\theta_D}({\rm vis} = 0|\boldsymbol{t}_j)$ is the probability that $\boldsymbol{t}_j$ is from the language space. On the contrary, the generator fools the discriminator from making correct predictions by minimizing the following objective:
\begin{equation}
\resizebox{0.9\hsize}{!}{$\begin{aligned}
\mathcal{L}_W(\boldsymbol{W_{align}}|\theta_D) = &-\frac{1}{m}\sum_{i = 1}^m\log{P_{\theta_D}({\rm vis} = 0|\boldsymbol{W_{{align}}}\boldsymbol{v}_i)} \\
                                                 &- \frac{1}{n}\sum_{j = 1}^n\log{P_{\theta_D}({\rm vis} = 1|\boldsymbol{t}_j)}
\end{aligned}$}
\label{f:3}
\end{equation}

\noindent\textbf{Coarse Calibration by Pseudo Structure Dictionary}. Since the adversarial training aligns regions and nouns regardless of their frequency, many regions with a small amount of data result in worse alignment. To calibrate the mapping, we build a structure dictionary from the roughly aligned region-noun embeddings after adversarial training. To ensure a high-quality dictionary, we consider the top frequent nouns to align, which are expected to have better alignment quality with more training samples. 
We regard the mutual nearest neighbors (MNN) \cite{lample2018word} between frequent noun embeddings and aligned region embeddings as pseudo region-noun pairs to construct the dictionary, \textit{i.e.}, pseudo structure dictionary. We use MNN instead of \textit{K}-nearest neighbors to avoid the hubness problem \cite{2014arXiv1412.6568D} (embeddings tending to be nearest neighbors of many embeddings).

To this end, we select the top $N$ ($N$=30) frequent nouns from $\mathcal{T}$, denoted as $\hat{\mathcal{T}}=\{\boldsymbol{\hat{t}}_j\}$, and then compute MNN of $\boldsymbol{\hat{t}}_j$ from the aligned region embeddings $\boldsymbol{W_{{align}}}\mathcal{V}$, denoted as $\hat{\mathcal{V}}=\{\boldsymbol{\hat{v}}_i\}$, to form the dictionary.  We utilize the Cross-Domain Similarity Local Scaling (CSLS) metric \cite{lample2018word} for MNN computation. We first select $L$ region embeddings $\{\boldsymbol{\hat{v}}_l\}\subset\hat{\mathcal{V}}$ with top $M$ cosine similarities for each $\boldsymbol{\hat{t}}_j$ in a training batch. For each $(\boldsymbol{\hat{v}}_l, \boldsymbol{\hat{t}}_j)$ pair, we compute MNN as $CSLS(\boldsymbol{\hat{v}}_l, \boldsymbol{\hat{t}}_j)=2\cos(\boldsymbol{\hat{v}}_l, \boldsymbol{\hat{t}}_j) - r_{\hat{\mathcal{T}}}(\boldsymbol{\hat{v}}_l)-r_{\hat{\mathcal{V}}}(\boldsymbol{\hat{t}}_j)$, where $r_{\mathcal{\hat{T}}}\left(\boldsymbol{\hat{v}}_l\right)=\frac{1}{K} \sum_{\boldsymbol{\hat{t}}_y \in \mathcal{N}_{\mathcal{\hat{T}}}\left(\boldsymbol{\hat{v}}_l\right)} \cos \left(\boldsymbol{\hat{v}}_l, \boldsymbol{\hat{t}}_y\right)$ computes the mean similarity between $\boldsymbol{\hat{v}}_l$ and its $K$ nearest neighbors in $\mathcal{\hat{T}}$, \textit{i.e.},  $\mathcal{N}_{\mathcal{\hat{T}}}\left(\boldsymbol{\hat{v}}_l\right)$. $r_{\hat{\mathcal{V}}}(\boldsymbol{\hat{t}}_j)$ shares similar operations but differs in similarity computation direction. We choose the nearest $\boldsymbol{\hat{v}}_l$ for each $\boldsymbol{\hat{t}}_j$ and form the dictionary with all the $\{\boldsymbol{\hat{v}}_l, \boldsymbol{\hat{t}}_j\}$ pairs. In practice, \cite{xing2015normalized} proved that the results are optimized by enforcing an orthogonality constraint on $\boldsymbol{W_{{align}}}$. Following \cite{xing2015normalized}, we regard Eq. \ref{f:Walign} as the Procrustes problem and provides a closed solution by the singular value decomposition (SVD) on the structure dictionary as $\boldsymbol{W_{align}} = UV^T, with\ U\Sigma{V^T}=SV\!D(\hat{\mathcal{T}}\hat{\mathcal{V}}^T)$.

\noindent\textbf{Accurate Refinement by Pseudo Semantic Dictionary.} 
Though the above two stages coarsely align vision and language spaces, they are difficult to align specific product characteristics accurately, \textit{e.g.}, mapping the ads image region to the distinctive description `black Nike T-shirt', since there lack of semantic annotations for fine-grained alignment. Therefore, we propose to automatically construct a semantic dictionary as supervision to refine further $\boldsymbol{W_{align}}$. Specifically, we first detect all the object regions \{$\boldsymbol{o}_i$\} with corresponding object labels \{$\boldsymbol{l}_i$\} in the ads images. Since the query generally describes product features by nouns, we select nouns \{$\boldsymbol{w}_j$\} in each query that match object labels in the relevant ads image satisfying $\boldsymbol{w}_j \in  \{\boldsymbol{l}_i\}$ to form pseudo region-noun pairs \{$\boldsymbol{o}_i$, $\boldsymbol{w}_j$\}. To represent $\boldsymbol{w}_j$ with comprehensive product features, we locate all $\boldsymbol{w}_j$ in both query and ads text and average their embeddings to represent $\boldsymbol{w}_j$. The region embeddings are the same as in Sec. 3.1. In this way, we build a dictionary with about 50K region-noun pairs and sample 20\% of them according to the distribution of nouns to form the semantic dictionary for refinement according to Eq. \ref{f:Walign} as Coarse Calibration.  

\subsection{3.3. AlignCMSS: Cross-Modal Sponsored Search Model with VALSE}
\label{ssec:search model}
Since VALSE is learned independently of the sponsored search model, we introduce the training strategy of incorporating $\boldsymbol{W_{align}}$ with a recently proposed VLP model, \textit{e.g.,} VinVL \cite{zhang2021vinvl}. We name our model as AlignCMSS. Following the design in VinVL, we also feed the object tags detected in the ads image together with the query, ads text, and ads image to the multi-modal encoder. As shown in Figure \ref{fig:model-architecture} (left), we directly replace the original linear mapping of the input layer in VinVL by VALSE to align the region embeddings to the language ones and fine-tune it with VinVL together. 
We initialize VinVL with its pre-trained weights and further pre-train AlignCMSS on the ads search dataset with two tasks: Masked Token Model (MTM) and Query-Ads Contrastive Learning.   

\noindent\textbf{Masked Token Model.} Similar to Mask Language Model \cite{devlin2019bert}, we randomly mask each input token $\boldsymbol{h}_i$ with the probability of 15\% from $m$ input tokens and replace the masked token with a special token \texttt{[MASK]}. 
The goal of MTM is to predict the masked tokens based on other tokens, denoted as $\hat{\boldsymbol{h}}_i$, by minimizing the negative log-likelihood:
\begin{equation}
\begin{aligned}
\mathcal{L}_{MTM} = -\mathbb{E}_{ \boldsymbol{h}_j\sim{\mathcal{D}}}\log{p(\boldsymbol{h}_i|{{\hat{\boldsymbol{h}}_i}})}
\label{f:6}
\end{aligned}
\end{equation}
\noindent\textbf{Query-Ads Contrastive Learning.} Following VinVL \cite{zhang2021vinvl}, we construct negative samples for each query-ads pair by replacing the query with a different query from the training set with a probability of 25\%. The training data with both positive and negative samples is denoted as $\mathcal{D}$. We feed the output embedding of \texttt{[CLS]} to a fully-connected layer as a classifier $f(\cdot)$ to predict whether the query matches the ads (\textit{c}=0) or not (\textit{c}=1). The contrastive loss is defined as:
\begin{equation}
\begin{aligned}
\mathcal{L}_{CL} = -\mathbb{E}_{(\boldsymbol{q}_i,\boldsymbol{a}_i;c)\sim{\mathcal{D}}}\log{p(c|f(\boldsymbol{q}_i, \boldsymbol{a}_i))}
\label{f:7}
\end{aligned}
\end{equation}

The final pre-training loss for AlignCMSS is defined as: $\mathcal{L} = \mathcal{L}_{MTM} + \mathcal{L}_{CL}$.

\noindent\textbf{Fine-tuning.}  We formulate the cross-modal sponsored search task as a binary classification problem as defined in \cite{DBLP:journals/corr/abs-2001-07966}. Given a query-ads pair ($\boldsymbol{q}_i, \boldsymbol{a}_i$) with ground-truth label $\boldsymbol{y}_i\in{\{0, 1\}}$, the output $\texttt{[CLS]}$ token $t_{(\boldsymbol{q}_i, \boldsymbol{a}_i)}$ of AlignCMSS is fed to a binary classifier to predict the relevance, \emph{i.e.} $\hat{\boldsymbol{y}_i} = p(t_{(\boldsymbol{q}_i, \boldsymbol{a}_i)})$. We construct negative samples by replacing queries $\boldsymbol{q}_i$ from positive samples, thereby making the model aware of the importance of user queries. The cross-entropy loss is defined as:
\begin{equation}
\begin{aligned}
\resizebox{0.9\hsize}{!}{$\mathcal{L}_{BCE} = -\mathbb{E}_{(\boldsymbol{q_i, a_i})\sim{\mathcal{L}}}[\boldsymbol{y}_i\log{\hat{\boldsymbol{y}_i}} + (1-\boldsymbol{y}_i)\log{(1-\hat{\boldsymbol{y}_i} )}]$}
\label{f:9}
\end{aligned}
\end{equation}
In the testing stage, the probability of $\hat{\boldsymbol{y}_i}$ is used to predict whether a given query-ads is relevant.

\section{4. Experiments}
\noindent\textbf{Dataset and Evaluation Metric.} We test AlignCMSS on a large commercial advertising dataset proposed in \cite{10.1145/3503161.3548226}, containing 480K query-ad pairs where 400K for training and 80K for testing. Queries are labeled with relevant and irrelevant ads. We evaluate the model by the Area under Reciever Operating Characteristic Curve (AUC) as in \cite{cortes2003auc}.

 \begin{table}[t]
\centering
\scalebox{1.1}
{\scriptsize
\renewcommand\arraystretch{1.1}
\setlength{\tabcolsep}{1.4mm}
\begin{tabular}{l|l|c}
\toprule
\multicolumn{1}{c|}{\textbf{Ads Input}}              & \multicolumn{1}{c|}{\textbf{Model}} & \textbf{AUC}   \\ \hline
\multicolumn{1}{c|}{Text}                   & BERT-base \cite{devlin2019bert}                & 82.15 \\ \hline
\multicolumn{1}{c|}{\multirow{3}{*}{Image}} & CLIP \cite{radford2021learning}                     & 81.16 \\
\multicolumn{1}{c|}{}                       & ViLT \cite{pmlr-v139-kim21k}                      & 82.23 \\
\multicolumn{1}{c|}{}                       & Unicoder-VL \cite{Li_Duan_Fang_Gong_Jiang_2020}              & 83.16 \\ \hline
\multirow{14}{*}{Multi-Modal}                & CLIP@(Qt-OtOi) \cite{10.1145/3503161.3548226}           & 81.98 \\
                                            & CLIP-MH@(Qt-OtOi) \cite{10.1145/3503161.3548226}        & 82.18 \\
                                            & ALBEF \cite{NEURIPS2021_50525975}       & 82.74 \\
                                            & BLIP \cite{pmlr-v162-li22n}       & 83.51 \\
                                            & VL-BERT \cite{su2020vl-bert}    & 86.27 \\  
                                            & OSCAR \cite{li2020oscar}                    & 87.45 \\
                                            & Unicoder-VL@(Qt-OtOi) \cite{10.1145/3503161.3548226}    & 87.90 \\                            
                                            & VinVL \cite{zhang2021vinvl}     & 88.56 \\
                                            & AdsCVLR \cite{10.1145/3503161.3548226}                  & 89.16 \\ \cline{2-3} 
                                            & VL-BERT+VALSE              & 90.13     \\
                                            & OSCAR+VALSE              & 90.62       \\
                                            & AlignCMSS (50\%)              & 90.46       \\
                                            & AlignCMSS (70\%)             & 91.02       \\
                                            & AlignCMSS (100\%)              & \textbf{91.73} \\ \bottomrule
\end{tabular}}
\caption{Comparison with state-of-the-art models.
}
\vspace{-10pt}
\label{tab:sota}
\end{table}

\noindent\textbf{Implementation Details.}
The vision space contains $1.5$ million region embeddings, and the language space has $1.6$ million word embeddings extracted from the dataset. VALSE is pre-trained by an SGD optimizer with $100$k iterations, where the mini-batch size is $1024$, and the learning rate is $0.1$. The input length of AlignCMSS is $152$. AlignCMSS is pre-trained by Adam optimizer with $180$k iterations, where the mini-batch size is $64$, and the dropout ratio is $0.1$. We set the learning rate to $1 \times 10^{-4}$. The model is fine-turned by $40$ epochs, and the learning rate is $2 \times 10^{-5}$. VALSE is trained for $16$ hours, while AlignCMSS is pre-trained for $216$ hours and fine-tuned for $24$ hours with $3$ NVIDIA V100 (32G). 

\subsection{4.1. Comparison with State-of-the-Art Methods}

 \begin{figure}[t]

    \centering
    \includegraphics[width=0.76\linewidth]{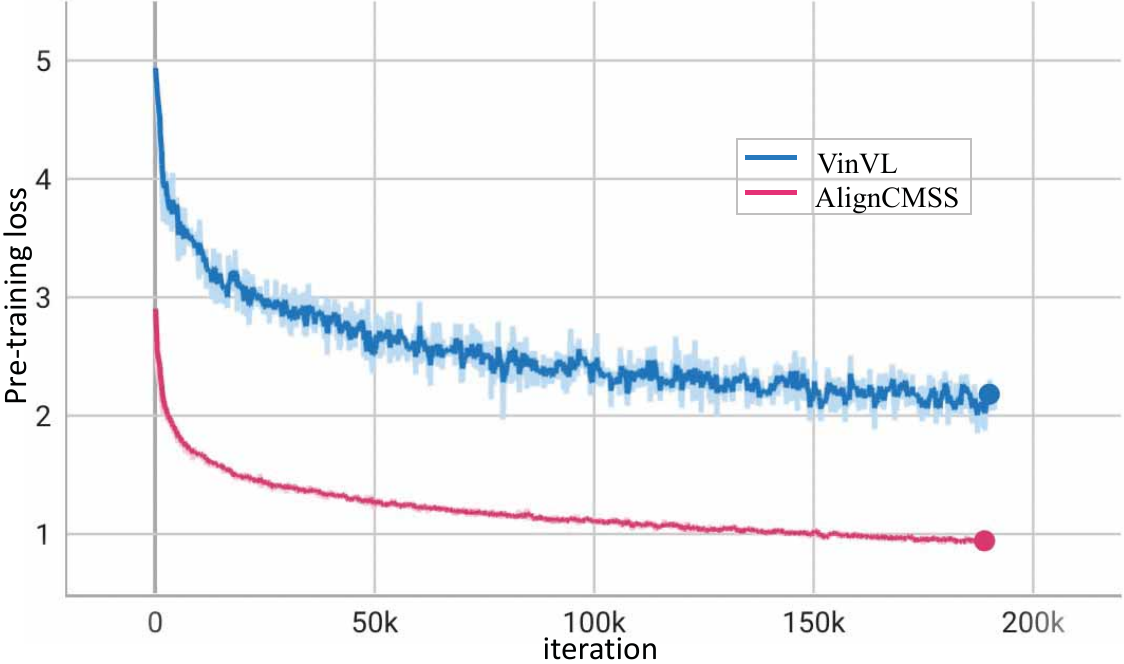}
    \caption{Comparison of the pre-training loss curves.
    }
    \vspace{-5pt}
    \label{fig:loss}

\end{figure}

Table \ref{tab:sota} shows the results of three kinds of state-of-the-art (SoTA) models: text-only model \cite{devlin2019bert},  image-only models \cite{radford2021learning, pmlr-v139-kim21k, Li_Duan_Fang_Gong_Jiang_2020}, and multi-modal models \cite{su2020vl-bert, li2020oscar, 10.1145/3503161.3548226, zhang2021vinvl, NEURIPS2021_50525975, pmlr-v162-li22n}. Besides VinVL, we also combined VALSE with another two widely compared VLP models, \textit{i.e} VL-BERT and OSCAR, in the last block to prove the generalization ability of our align-and-match strategy. We also pre-trained and fine-tuned AlignCMSS with 50\%, 70\%, and 100\% training data to prove its effectiveness.

The results show that AlignCMSS consistently outperforms all existing approaches, achieving a new SoTA performance that significantly outperforms the SoTA model AdsCVLR by 2.57\%. VALSE also significantly improves VL-BERT and OSCAR by 3.86\% and 3.17\%, respectively. Moreover, AlignCMSS outperforms AdsCVLR by 1.3\% with only 50\% of the training data, proving that unifying heterogeneous inputs in language space through VALSE significantly enhances the relevance of measuring the VLP model's ability.
Besides the advantages of using less training data, AlignCMSS achieves a faster and more stable learning process than VinVL, as shown by the pre-training loss curves in Figure \ref{fig:loss}, because VALSE effectively bridges the semantic gap between ads images and text and alleviates the burden of cross-modal correlation learning.

\begin{table}[t]
\centering
\scalebox{1.08}
{\scriptsize
\begin{tabular}{llc}
\toprule
         & \multicolumn{1}{l|}{\textbf{Method}}                & \textbf{AUC}   \\ \hline
1.       & \multicolumn{1}{l|}{AlignCMSS (full model)}                     & \textbf{91.73} \\ \hline
\multicolumn{2}{l|}{\textbf{Ablation of Three Stage Alignment}}         &                \\
2.       & \multicolumn{1}{l|}{w/o Accurate Refinement}          & 90.34          \\
3.       & \multicolumn{1}{l|}{w/o Coarse Calibration}       & 91.16          \\
4.       & \multicolumn{1}{l|}{w/o   Distribution Alignment \& Coarse Calibration}         & 90.65          \\
5.      & \multicolumn{1}{l|}{w/o    Coarse Calibration \& Accurate Refinement}        & 89.97          \\
6.      & \multicolumn{1}{l|}{w/o    VALSE Alignment}        & 89.51          \\
\multicolumn{2}{l|}{\textbf{Ablation of Training Strategies}} &                \\
7.       & \multicolumn{1}{l|}{w/o $\mathcal{L}_{MTM}$}                       &  91.37              \\
8.       & \multicolumn{1}{l|}{w/o $\mathcal{L}_{CL}$}                        &  91.46              \\
9.       & \multicolumn{1}{l|}{w/o $\mathcal{L}_{MTM}$ \& $\mathcal{L}_{CL}$}                      &     91.13           \\
10.       & \multicolumn{1}{l|}{w/o Fine-tuning}                      &      91.04          \\
\multicolumn{2}{l|}{\textbf{Ablation of Pre-trained Knowledge in VinVL}} &                \\
11.       & \multicolumn{1}{l|}{w/o VinVL pre-training}                       &    90.14         \\
\multicolumn{2}{l|}{\textbf{Ablation of Object Tags in VinVL}} &                \\
12.       & \multicolumn{1}{l|}{w/o Object Tags}                       &    91.55         \\
\bottomrule
\end{tabular}}
\caption{Ablation of key components in AlignCMSS.}
\vspace{-5pt}
\label{tab:ablation}
\end{table}

\subsection{4.2. Ablation Study}
In Table \ref{tab:ablation}, we evaluate the effectiveness of three alignment stages, training losses, and pre-trained knowledge in AlignCMSS. (1) In models ‘2-6’, we observe that each alignment stage is indispensable for the performance boost. When removing accurate refinement (model ‘2’), the AUC score decreases by 1.39\%, which is more significant than removing coarse calibration. It indicates that weak supervision is essential for bridging the semantic gap across different modalities. Only weakly-supervised alignment (model ‘4’) or only unsupervised adversarial training (model `5') leads to further performance decrease. Without VALSE (model `6'), AUC decreases obviously by 2.22\%. It proves that the three stages of VALSE work jointly to achieve the maximum boost. (2) In models ‘7-10’, we conclude that the masked token model and contrastive learning are effective pre-training strategies, while fine-tuning is also necessary to achieve the best performance. Combining with other pre-training tasks, such as image classification in AdsCVLR \cite{10.1145/3503161.3548226}, can be studied in future work. (3) Without the pre-trained weights in VinVL (model ‘11’),  the AUC score drops by 1.59\%, indicating that implicit vision-language correlations learnt by pre-training benefit relevance learning. 
It is worth noticing that the AUC score of model ‘11’ is still 0.63\% higher than model `6’, proving that the explicit vision and language alignment by VALSE is more beneficial for relevance modeling than implicit alignment by pre-training. (4) Without the object tags in AlignCMSS (model ‘12’), the AUC score only slightly decreases by 0.18\%, which indicates that VALSE achieves the major contribution for alignment instead of tags. 

\begin{table}[t]
\centering
\scalebox{1.1}
{\scriptsize
\renewcommand\arraystretch{1.1}
\begin{tabular}{ll|c}
\toprule
                          & \textbf{Model}                                           & \textbf{AUC} \\ \hline
\multicolumn{2}{l|}{\textbf{Semantic Dictionary Size}} &                                      \\
\multicolumn{1}{l}{1.}  & \multicolumn{1}{l|}{Semantic dictionary size-100\%}                    &    91.82          \\
\multicolumn{1}{l}{2.}  & \multicolumn{1}{l|}{Semantic dictionary size-20\%}                     &    91.73          \\
\multicolumn{1}{l}{3.}  & \multicolumn{1}{l|}{Semantic dictionary size-2\%}                      &    91.20          \\
\multicolumn{1}{l}{4.}  & \multicolumn{1}{l|}{Semantic dictionary size-0.2\%}                    &    90.96          \\
\multicolumn{1}{l}{5.}  & \multicolumn{1}{l|}{Semantic dictionary size-0\%}                        &    90.34          \\
\multicolumn{2}{l|}{\textbf{Noun Selection in the Semantic Dictionary}} &                               \\
\multicolumn{1}{l}{6.}  & \multicolumn{1}{l|}{Noun-head-20\%}                &    91.16          \\
\multicolumn{1}{l}{7.}  & \multicolumn{1}{l|}{Noun-tail-20\%}                &    90.86          \\
\multicolumn{1}{l}{8.} & \multicolumn{1}{l|}{Noun-random-20\%}              &    91.61          \\ 
\multicolumn{2}{l|}{\textbf{Unsupervised Alignment Method}} &                                       \\
\multicolumn{1}{l}{9.}  & \multicolumn{1}{l|}{Wasserstein Procrustes}              &    90.15          \\
\multicolumn{1}{l}{10.}  & \multicolumn{1}{l|}{Accurate refinement  \& Dictionary size-100\%} &    91.18          \\ 
\multicolumn{2}{l|}{\textbf{Language Embedding}} &                               \\
\multicolumn{1}{l}{11.}  & \multicolumn{1}{l|}{BERT noun token Embedding}                    &   90.79          \\
\multicolumn{1}{l}{12.} & \multicolumn{1}{l|}{Single noun Embedding}              &    91.22          \\ 
\bottomrule
\end{tabular}}
\caption{Analysis of key components in VALSE.}
\vspace{-5pt}
\label{tab:ablation3}
\end{table}

\subsection{4.3. Key Component Analysis in VALSE}
\label{ssec:align analysis}

We conduct extensive experiments on alternative methods of key components in VALSE to prove the advantages of the selected approach in Table \ref{tab:ablation3}.  
(1) In models ‘1-5’, we evaluate the effect of semantic dictionary size on the search performance. 
We range the dictionary size from 100\% to 0\% (\textit{i.e.} without the accurate refinement stage) and conclude that more pseudo semantic supervision achieves better performance. Model `2' (full model) with only 20\% pseudo region-noun pairs obtains comparable performance with the fully supervised counterpart (model `1'), which proves the effectiveness of VALSE with relatively low training resources. (2) In models ‘6-8’, we assess the influence of noun classes in the semantic dictionary on performance. We build four dictionaries with the same data size (20\%) in different ways: head 30 frequent nouns (model `6'), tail 30 frequent nouns (model `7'), and randomly selected nouns (model `8'). The AUC score of models `6’ and ‘8’ are higher, which indicates that the more consistent between the distribution of supervised data and aligned data, the better alignment is achieved. (3) In model ‘9’, we replace VALSE with another typical unsupervised distribution alignment method Wasserstein Procrustes \cite{grave2019unsupervised}. The performance decreases by 1.58\% compared with AlignCMSS. In model `10', we only retain the weakly-supervised stage with 100\% training data, which achieves inferior performance compared to the full model with the unsupervised alignment stages. In summary, the unsupervised alignment strategy in VALSE is effective and complementary to the weakly-supervised stage.
(4) In models ‘11-12’, both the original BERT noun token embeddings and the single noun embeddings (without averaging) are inferior to the averaged noun embeddings output by VinVL, which contain richer co-occurrence information for structure alignment.

\begin{figure}[t]
    \centering
    \includegraphics[width=1.0\linewidth]{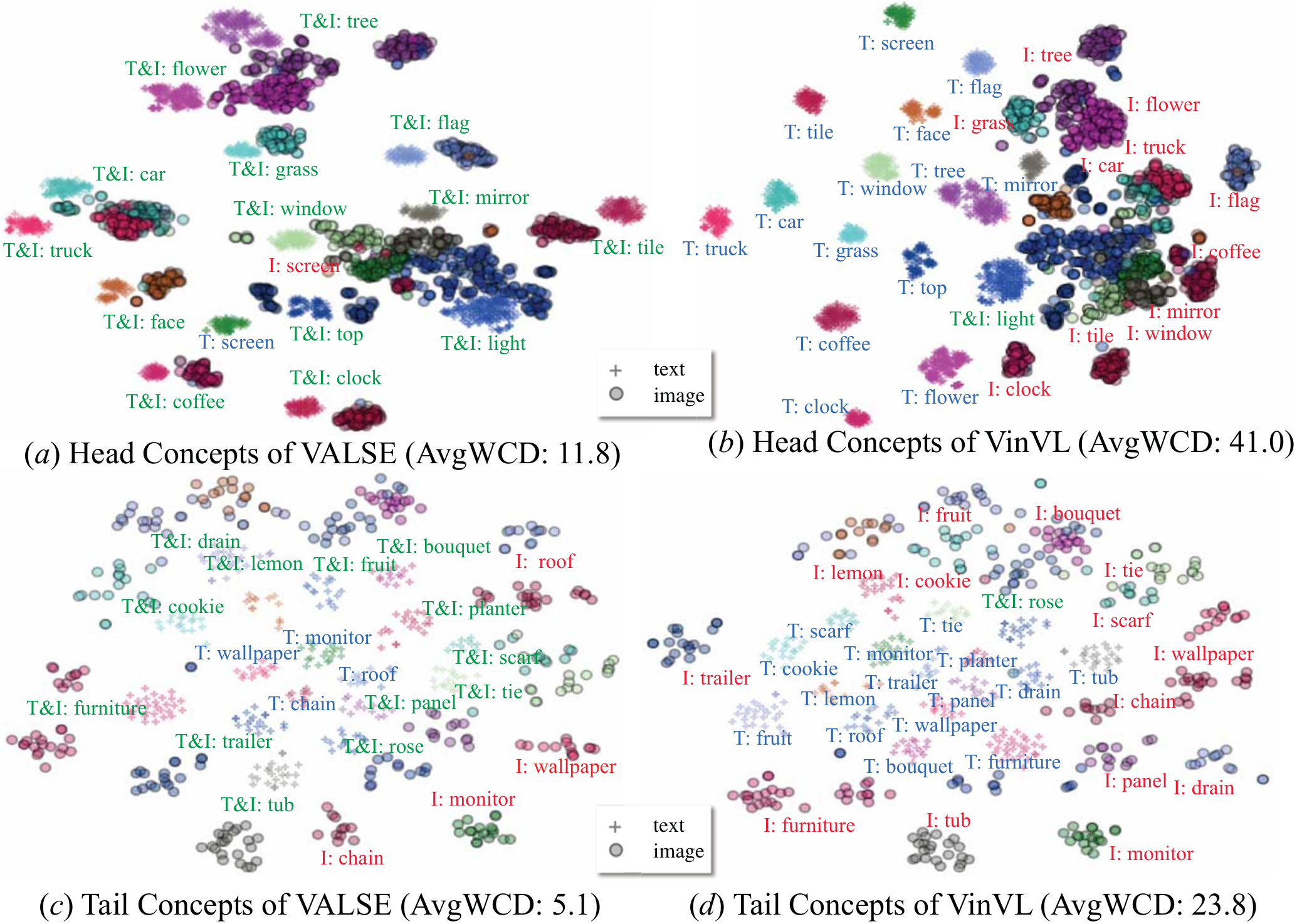}
    \caption{T-SNE visualization of the aligned region and noun embeddings.  ‘Cross’ and ‘\textcolor{blue}{T}’ indicate nouns, ‘circle’ and ‘\textcolor{red}{I}’ mean aligned regions, and ‘\textcolor{green}{T\&I}’ means well-aligned regions and nouns.  Best viewed in color with 300\% zoom.}
    \vspace{-10pt}
    \label{fig:t-SNE}
\end{figure}

\subsection{4.5. Alignment Analysis}
We conduct experiments to prove the alignment quality between regions and nouns and VALSE's generalization ability. Please refer to the supplementary for a detailed analysis.

\noindent{\textbf{(1) Alignment Quality between Regions and Words.}}
We use t-SNE \cite{van2008visualizing} to visualize both head and tail nouns and aligned region embeddings. We compare VALSE with the outputs of VinVL in Figure \ref{fig:t-SNE}. VALSE aligns region embeddings with related nouns much closer than VinVL over both head and tail concepts. Moreover, we obtain the same conclusion from the quantitative evaluation by Average Within-Cluster Distance (AvgWCD) \cite{edwards1965method}. 
The smaller AvgWCD means the better aligned quality. VALSE consistently outperforms VinVL on this metric.

\noindent{\textbf{(2) Alignment Generalization Ability.}} VALSE is a generic alignment approach and can be applied to other cross-modal scenarios.
We case-study another cross-modal retrieval task on the MSCOCO dataset \cite{10.1007/978-3-319-10602-1_48} to prove its generalization ability. Table \ref{tab:coco} shows the retrieval results on 1K and 5K test sets. Compared to VinVL, AlignCMSS achieves a consistent performance boost on text retrieval and image retrieval tasks over all the metrics. 

\begin{table}[t]
\centering
\setlength{\abovecaptionskip}{5pt}
\setlength{\belowcaptionskip}{-10pt}
\setlength{\tabcolsep}{0.91mm}
\scalebox{0.93}
{\scriptsize
\renewcommand\arraystretch{1.1}
\begin{tabular}{l|cc|cc}
\toprule
\multirow{3}{*}{Model} & \multicolumn{2}{c|}{1K Test Set}          & \multicolumn{2}{c}{5K Test Set}         \\
                       & Text Retrieval      & Image Retrieval    & Text Retrieval     & Image Retrieval    \\ \cline{2-5} 
                       & R@1 / 5 / 10       & R@1 / 5 / 10      & R@1 / 5 / 10      & R@1 / 5 / 10      \\ \hline
VinVL                  & 89.8 /  98.8 / 99.7 & 78.2 / 95.6 / 98.0 & 74.6 / 92.6 / 96.3 & 58.1 / 83.2 / 90.1 \\
AlignCMSS                  & \textbf{91.2 /  99.4 / 99.8} & \textbf{80.1 / 97.3 / 99.2} &  \textbf{77.4 / 94.1 / 97.0}    & \textbf{60.4 / 84.3 / 90.7}    \\
$\Delta\uparrow$       & \textbf{\hphantom{1}1.4 /\hphantom{1} 0.6 / \hphantom{1}0.1}     & \textbf{\hphantom{1}1.9 / \hphantom{1}1.7 / \hphantom{1}1.2}    & \textbf{\hphantom{1}2.8 / \hphantom{1}1.5 / \hphantom{1}0.7}                  & \textbf{\hphantom{1}2.3 / \hphantom{1}1.1 / \hphantom{1}0.6}           \\ \bottomrule
\end{tabular}}
\vspace{1pt}
\caption{Results of image-text retrieval on MSCOCO.}
\label{tab:coco}
\end{table}

\section{5. Conclusion}

In this paper, we propose a novel alignment method VALSE to align ads regions to ads texts based on their structure consistency. Combined VALSE with VinVL, we propose a new cross-modal sponsored search model AlignCMSS with state-of-the-art performance on a large commercial dataset. Extensive experiments also prove its advantages of high alignment quality, efficient pre-training, and generalization ability. The application of VALSE in other cross-modal scenarios will be explored in our future work.

\bibliography{aaai24}

\end{document}